\documentclass{article} 
\usepackage{arxiv,times}

\usepackage{amsmath,amsfonts,bm}

\def\eqref#1{equation~\ref{#1}}

\def\1{\bm{1}}

\DeclareMathAlphabet{\mathsfit}{\encodingdefault}{\sfdefault}{m}{sl}
\SetMathAlphabet{\mathsfit}{bold}{\encodingdefault}{\sfdefault}{bx}{n}

\usepackage{hyperref}
\usepackage{url}

\usepackage[table]{xcolor}
\usepackage{booktabs}
\usepackage{multirow}
\usepackage{graphicx}
\usepackage{xcolor}
\usepackage{float}
\usepackage{dblfloatfix}
\usepackage{algorithm}
\usepackage{algpseudocode}
\usepackage{amssymb}

\usepackage{wrapfig}
\usepackage{caption}
\title{HGA-Quant: Heterogeneity-Aware Quantization for AutoRegressive Visual Generation Models}

\author{%
 Xinmeng Liu$^{1}$\thanks{\scriptsize Equal contribution}~,\enspace Xianglong Yan$^{1}$\footnotemark[1]~,\enspace Kaicheng Yang$^{1}$,\enspace Weisheng Dong$^{2}$,\enspace Yulun Zhang$^{1}$\thanks{\scriptsize Corresponding author: Yulun Zhang, yulun100@gmail.com}~\enspace \\ \textsuperscript{1}Shanghai Jiao Tong University,\enspace \textsuperscript{2}Xi'an University of Electronic Science and Technology\enspace 
}

\iclrfinalcopy 
\begin{document}

\maketitle

\vspace{-10mm}

\begin{figure}[h]
    \centering
    \includegraphics[width=\linewidth]{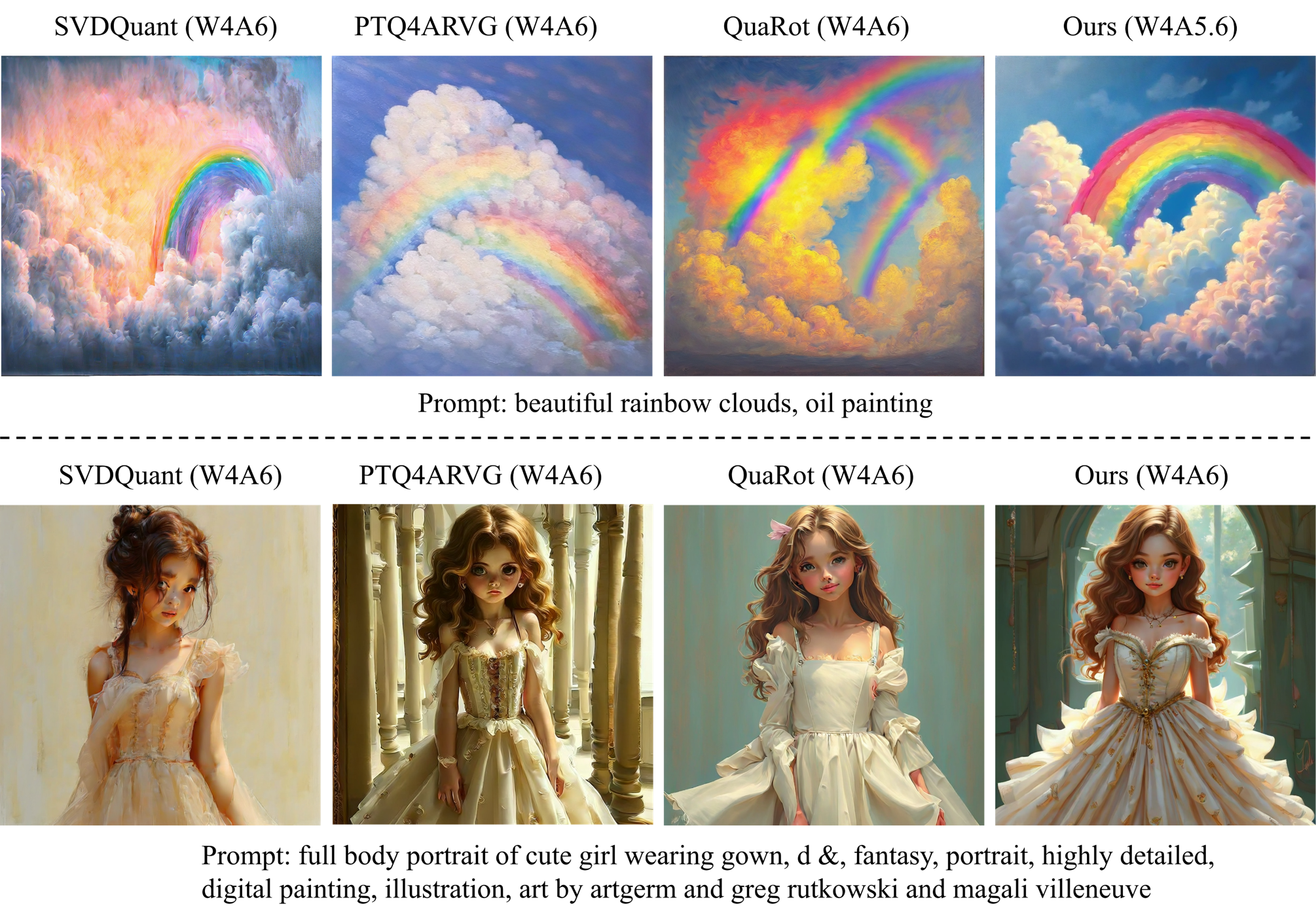}
    \label{fig:display_p1}
\end{figure}

\vspace{-8mm}

\begin{abstract}
Recently, AutoRegressive Visual Generation (ARVG) models have emerged as a competitive paradigm for high-quality image synthesis. However, their substantial computational and memory demands make them difficult to deploy on resource-constrained edge devices. Post-Training Quantization (PTQ) provides a promising solution to reduce the computational and memory costs of ARVG models, yet existing methods often suffer from substantial image quality degradation under low-bit quantization. In this paper, we propose a training-free PTQ framework for ARVG, consisting of Tail-Aware Bounded-init Grid Refinement (T-BGR) and Coarse-to-Fine Activation Precision Scheduling (CAPS). T-BGR introduces a tail-aware reconstruction objective for weight quantizer calibration, assigning greater importance to distribution-tail coefficients and preserving fine-grained information without introducing additional parameters or high-precision branches. CAPS performs coarse-to-fine activation precision scheduling by jointly considering the importance and computational cost of different generation stages, enabling efficient mixed-precision allocation under a given bit budget. Extensive experiments on multiple ARVG backbones demonstrate that our framework consistently improves robustness under aggressive low-bit quantization. Notably, on Infinity, our framework achieves near full-precision image generation quality at an effective W4A5.6 activation precision, while existing methods at W4A6 still exhibit noticeable quality degradation or inconsistent performance. Code is available at \href{https://github.com/lxm-lc/HGA-Quant}{here}.
\end{abstract}

\section{Introduction}

Autoregressive visual generation (ARVG) models have recently emerged as a competitive paradigm for high-quality image synthesis. A representative line of work, Visual Autoregressive Modeling (VAR), reformulates image generation as next-scale prediction and constructs images from coarse to fine~\citep{VAR}. Subsequent models have explored residual refinement, flexible scale schedules, and bitwise token prediction, as exemplified by HART~\citep{HART}, FlexVAR~\citep{FlexVAR}, and Infinity~\citep{Infinity}. These models provide a compelling alternative to diffusion-based generation, but their large Transformer backbones and repeated token prediction still impose substantial computational and memory demands. Such costs become particularly restrictive when ARVG models are deployed outside large-scale data-center settings.

Post-Training Quantization (PTQ) is an attractive way to reduce the cost of pretrained generative models without retraining them. By representing weights and activations with low-bit integers, PTQ can reduce the numerical precision required by the dominant linear operations while retaining the original model structure. However, directly applying generic PTQ methods to ARVG models often causes severe image-quality degradation at aggressive bit-widths. Recent studies have identified channel-wise outliers, token-wise activation dynamics, and calibration-distribution mismatch as important sources of this difficulty~\citep{PTQ4ARVG}. Other methods mitigate quantization errors through equivalent transformations~\citep{SmoothQuant}, rotations~\citep{QuaRot}, or auxiliary low-rank compensation~\citep{SVDQuant}. Despite these advances, the quality of visual autoregressive generation can still collapse when both weights and activations are quantized to low precision, indicating that the non-uniform structure of ARVG quantization errors requires more specialized treatment~\citep{Shift-and-Sum}.

We identify two complementary forms of heterogeneity that are especially relevant to low-bit ARVG quantization. First, weight coefficients are distributed unevenly across a large central mass and a much smaller tail region. Under an unweighted reconstruction loss, the central mass of a channel usually dominates the calibration objective, while tail coefficients contribute relatively little despite their larger magnitudes. Consequently, a calibration objective that minimizes average coefficient error may favor central-mass fidelity while leaving the tail reconstruction trade-off insufficiently controlled. Existing outlier-handling approaches often redistribute the quantization difficulty or preserve outliers through additional transformations or compensation paths. While these designs can improve numerical reconstruction, they do not directly expose how the tail-versus-central reconstruction trade-off should affect the calibration of the scale and zero-point.

Second, the effect of activation quantization can vary along the ARVG generation trajectory. In the coarse-to-fine process of VAR-style models, early stages establish the global structural context on which later stages condition, while later stages operate on denser token grids and mainly refine local content~\citep{VAR}. Uniform activation precision therefore ignores the position-dependent sensitivity of the generation trajectory as well as its computational cost. Protecting all stages equally can allocate precision to late refinements, whereas assigning precision only according to token count may overlook the different consequences of perturbing early and late stages. These observations motivate a quantization framework that answers two different questions: which values in a weight distribution deserve greater reconstruction priority, and when along the generation trajectory should higher activation precision be used?

To address these challenges, we propose Heterogeneity-Aware Quantization for AutoRegressive Visual Generation models (HGA-Quant), a training-free PTQ framework composed of two complementary components. Tail-Aware Bounded-init Grid Refinement (T-BGR) extends bounded grid refinement with a tail-aware reconstruction objective. It identifies distribution-tail coefficients using channel statistics and assigns them larger weights during the search for the quantization interval, scale, and zero point. T-BGR keeps the standard quantized-weight representation while directing the calibration objective toward tail reconstruction. Coarse-to-Fine Activation Precision Scheduling (CAPS) exploits the generation-order prior of ARVG models. It estimates a stage-wise floating-point operations (FLOPs) proxy for the targeted linear projections and allocates the available high-precision budget to an early prefix of coarse stages, while using lower precision for the remaining fine stages. Thus, T-BGR targets non-uniform reconstruction priorities within the weight distribution, whereas CAPS targets non-uniform sensitivity along the generation trajectory.

We evaluate HGA-Quant on four ARVG backbones, including Infinity-2B~\citep{Infinity}, HART~\citep{HART}, VAR-d30~\citep{VAR}, and FlexVAR-d24~\citep{FlexVAR}, using $W_{b}A_{a}$ to denote $b$-bit weights and $a$-bit activations, under W8A8, W4A8, W4A6, and W4A4 configurations. Across models and evaluation metrics, T-BGR remains competitive at higher precision and becomes increasingly robust relative to existing baselines as the bit-width decreases. On Infinity-2B, the stage-wise analysis in Figure~\ref{fig:caps_stage_sensitivity} shows that the same A4 perturbation is substantially more damaging at early stages than at late stages. Accordingly, assigning W4A8 to the first 11 stages and W4A4 to the final two stages yields a FLOPs-weighted effective activation precision of 5.5711 bits, denoted W4A5.6 below, while maintaining performance comparable to uniform W4A6. The main tables report both uniform T-BGR (\emph{Ours}) and the CAPS variant (\emph{Ours (CAPS)}) where evaluated; the stage-wise and budgeted allocation studies further isolate the role of precision placement.

\begin{figure}[t]
    \vspace{-10mm}
    \centering
    \includegraphics[width=\linewidth]{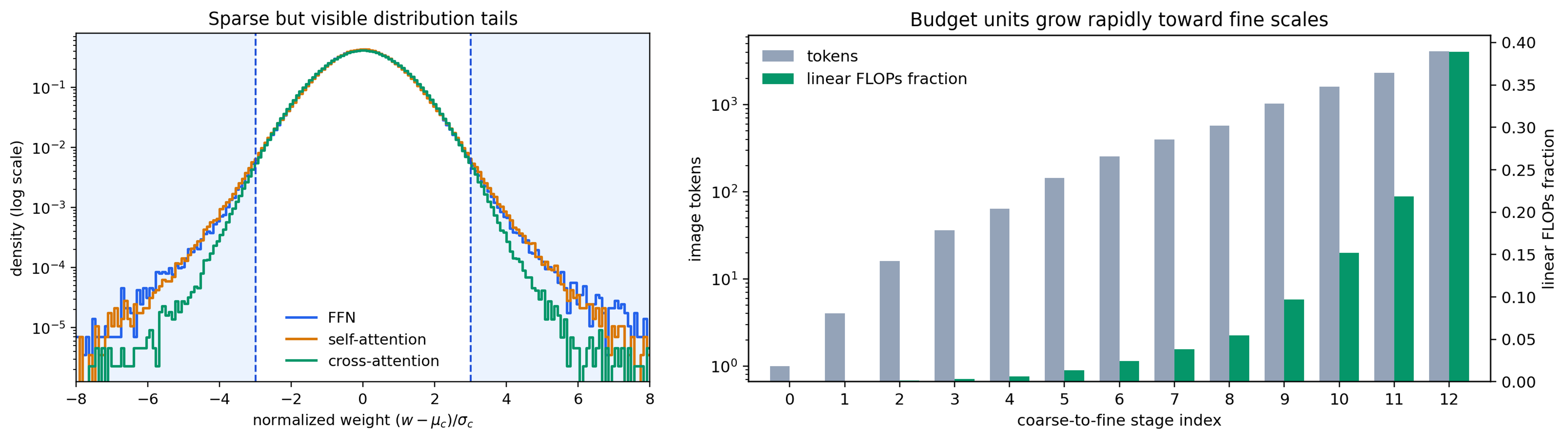}
    \vspace{-8mm}
    \caption{Motivation for HGA-Quant. Left: most weights concentrate near the central mass, but a non-negligible fraction lies in the tails, motivating tail-aware calibration. Right: token counts and linear-layer FLOPs grow sharply in later stages, motivating coarse-to-fine precision scheduling.}
    \vspace{-7mm}
    \label{fig:motivation}
\end{figure}

\section{Related Works}

\subsection{AutoRegressive Visual Generation models}

Recent ARVG models have emerged as a competitive paradigm for image synthesis. Representative approaches explore diverse autoregressive formulations, including scale-wise prediction~\citep{VAR}, continuous-token modeling~\citep{MAR}, randomized ordering~\citep{RAR}, parallelized generation~\citep{PAR}, and hybrid tokenization~\citep{HART}. VAR~\citep{VAR} is a foundational work that reformulates image generation as coarse-to-fine next-scale prediction, achieving strong image quality, inference efficiency, data efficiency, and scalability.
HART~\citep{HART} combines discrete autoregression with lightweight residual diffusion to improve reconstruction quality and enable efficient high-resolution generation. FlexVAR~\citep{FlexVAR} replaces residual prediction with ground-truth prediction at each scale, improving flexibility in resolution, aspect ratio, and inference steps. Infinity~\citep{Infinity} further scales ARVG through bitwise token prediction and an effectively enlarged token vocabulary. Its extensions have also explored unified image-video generation and image editing~\citep{InfinityStar,EditInfinity}. Alongside architectural advances, recent studies have investigated efficient decoding~\citep{CoDe,ZipAR}, attention compression~\citep{LiteVAR}, and low-bit quantization~\citep{LiteVAR,FPQVAR} to reduce the computational and memory costs of ARVG models.

\subsection{Post-Training Quantization}

Quantization is a standard compression technique for large generative models. It reduces model size, memory traffic, and inference latency. PTQ enables efficient deployment without retraining. GPTQ~\citep{GPTQ} uses Hessian-guided error compensation to minimize quantization error. AWQ~\citep{AWQ} protects salient weights based on activation statistics. Methods based on scale redistribution and equivalent transformations~\citep{OmniQuant,SmoothQuant,OutlierSuppressionPlus} mitigate activation outliers by shifting or redistributing quantization difficulty across weights and activations. Rotation-based methods~\citep{QuaRot,DVDQuant} further transform the representation space to disperse outliers across dimensions, facilitating low-bit quantization. SVDQuant~\citep{SVDQuant} and Q-DiT4SR~\citep{Q-DiT4SR} use a high-precision low-rank branch to compensate for quantization errors after outlier transformation. PTQ4ARVG~\citep{PTQ4ARVG} specifically studies PTQ for autoregressive visual generation, addressing channel-wise outliers, token-wise activation dynamics, and sample-wise distribution mismatch. Some recent works~\citep{Shift-and-Sum,DBLP:conf/icpr/VezzaliBGBL26} have begun to explore PTQ and quantization specifically for visual autoregressive models, targeting different aspects of efficient deployment. Despite these advances, significant image quality degradation remains under aggressive quantization configurations such as 4-bit weights and 6-bit activations in visual autoregressive models.

\section{Method}
\label{sec:method}

\begin{figure}[t]
    \vspace{-10mm}
    \centering
    \includegraphics[width=\linewidth]{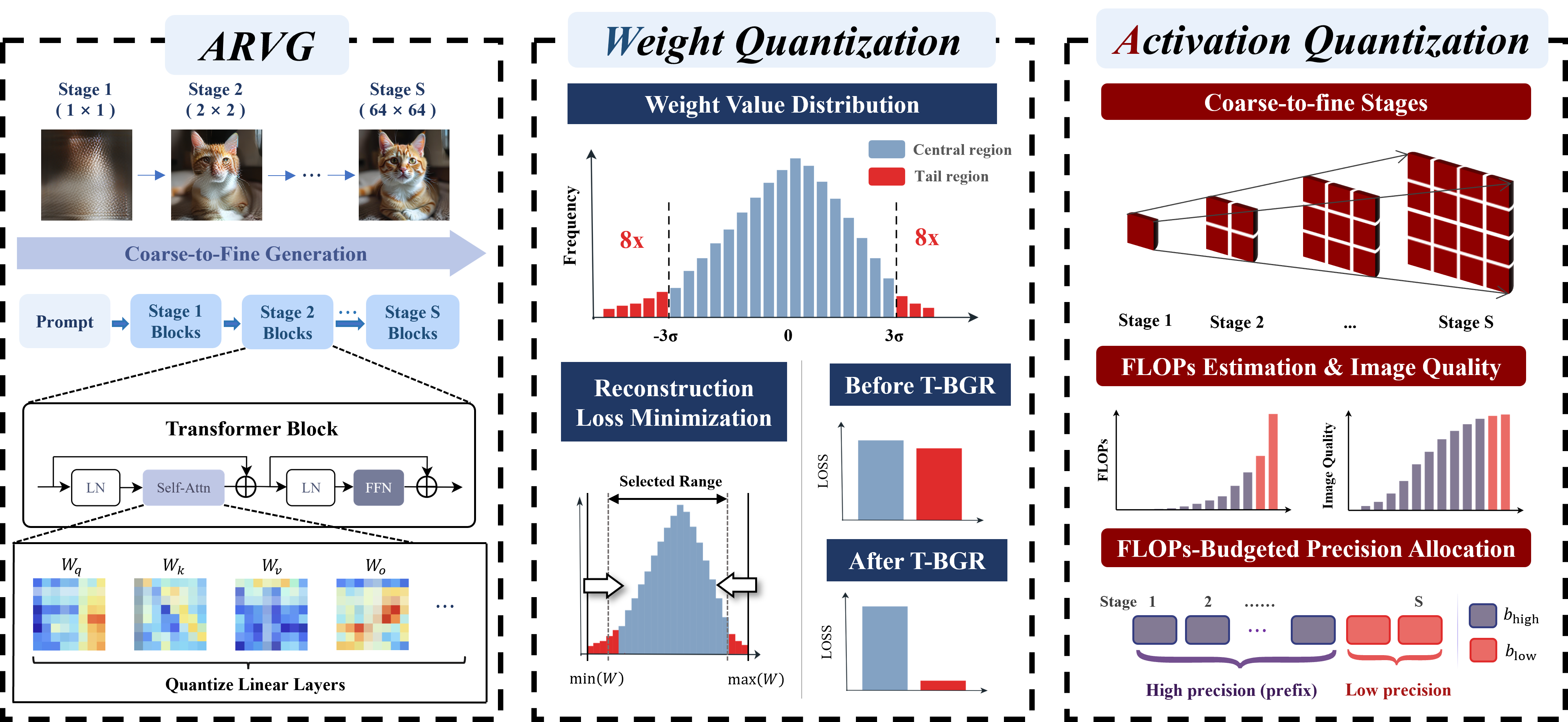}
    \vspace{-6mm}
    \caption{Overview of HGA-Quant. T-BGR performs tail-aware calibration for robust low-bit quantization, while CAPS assigns stage-wise precision in a coarse-to-fine manner for high-quality and efficient ARVG generation.}
    \label{fig:main_method}
    
\end{figure}

\subsection{Preliminaries}
\label{sec:method_overview}

Autoregressive image generators such as Infinity synthesize an image through a sequence of coarse-to-fine stages. Let $\mathcal{S}=\{\mathbf{p}_i\}_{i=0}^{S-1}$ denote the scale schedule, where $\mathbf{p}_i=(p_{i,1},\ldots,p_{i,D})$ specifies the token grid at stage $i$. The number of image tokens processed at this stage is $T_i=\prod_{d=1}^{D}p_{i,d}$. This generation process exposes two forms of non-uniformity that are not captured by conventional uniform post-training quantization (PTQ): different regions of a weight distribution can have different effects on the reconstructed output, and errors introduced at different generation stages need not propagate equally to the final image. We address these two sources of heterogeneity with T-BGR for tail-aware weight-quantizer calibration and CAPS for activation precision scheduling.

Post-training quantization converts a pretrained full-precision model to a low-bit model by determining quantization parameters from the weights and, when applicable, a small calibration set, without updating model parameters. For a scalar $x$, an unsigned $b$-bit asymmetric affine quantizer is written as
\begin{equation}
    \mathcal{Q}_{b}(x;s,z)
    = s\left(\operatorname{clip}\left(\left\lfloor \frac{x}{s}\right\rceil+z,0,2^b-1\right)-z\right),
    \label{eq:affine_quantizer}
\end{equation}
Here, $s$ is the scale and $z$ is the zero point. The corresponding integer code is $q=\operatorname{clip}(\lfloor x/s\rceil+z,0,2^b-1)$, and the expression in Eq.~\eqref{eq:affine_quantizer} is its dequantized value.

\subsection{Tail-Aware Bounded-init Grid Refinement}
\label{sec:tbgr}

\subsubsection{Motivation}

Bounded-init Grid Refinement (BGR) searches candidate quantization intervals and refines the corresponding affine scale and zero point to reduce weight reconstruction error~\citep{DVDQuant}. Its conventional reconstruction objective weights all coefficients equally. Under this objective, the large central mass dominates the total error, while a small number of tail coefficients contributes relatively little to the loss despite their larger magnitudes. Therefore, preserving these coefficients might be important for retaining fine-grained responses under aggressive quantization.

We therefore propose \emph{Tail-Aware Bounded-init Grid Refinement} (T-BGR), a tail-sensitive BGR extension. It modifies the calibration objective so that distribution-tail errors carry greater weight when selecting the quantization interval, scale, and zero point. This preserves the BGR deployment form while biasing its search toward a controlled tail-versus-central reconstruction trade-off.

\subsubsection{Tail-Aware Reconstruction Objective}

Consider one output channel $\mathbf{w}_c\in\mathbb{R}^{C_{\mathrm{in}}}$ of a weight matrix, where $w_{c,j}$ denotes its $j$-th weight coefficient. We compute its channel-wise mean and standard deviation as
\begin{equation}
    \mu_c=\frac{1}{C_{\mathrm{in}}}\sum_j w_{c,j},
    \qquad
    \sigma_c=\sqrt{\frac{1}{C_{\mathrm{in}}}\sum_j\left(w_{c,j}-\mu_c\right)^2}.
    \label{eq:tbgr_statistics}
\end{equation}
We identify a tail coefficient using a $\kappa\sigma$ criterion,
\begin{equation}
    m_{c,j}=\mathbf{1}\!\left[\left|w_{c,j}-\mu_c\right|>\kappa\sigma_c\right],
    \label{eq:tbgr_mask}
\end{equation}
and assign it the reconstruction weight
\begin{equation}
    \rho_{c,j}=1+(\gamma-1)m_{c,j}.
    \label{eq:tbgr_weight}
\end{equation}

Here, $\kappa$ controls the statistical tail boundary and $\gamma$ controls the relative priority assigned to tail coefficients. A larger $\gamma$ increases the influence of tail reconstruction on the selected quantization parameters, while a smaller $\kappa$ identifies a larger portion of the channel as tail.

Let $\hat{w}_{c,j}=\mathcal{Q}_{b_w}(w_{c,j};s_c,z_c)$ be the dequantized value of $w_{c,j}$, where $b_w$ denotes the weight bit-width. T-BGR minimizes the normalized tail-aware reconstruction error
\begin{equation}
    \mathcal{L}_{\mathrm{T\text{-}BGR}}^{(c)}(s_c,z_c)
    =
    \frac{\sum_j \rho_{c,j}\left(\hat{w}_{c,j}-w_{c,j}\right)^2}
    {\sum_j\rho_{c,j}}.
    \label{eq:tbgr_loss}
\end{equation}
When $\gamma=1$, Eq.~\eqref{eq:tbgr_loss} reduces to the uniformly weighted reconstruction criterion used by the original BGR calibration. For $\gamma>1$, the selected quantization parameters place greater emphasis on the reconstruction of tail coefficients, while the normalization keeps the loss scale comparable across channels with different tail ratios.

\subsubsection{Weighted Grid Refinement}

T-BGR retains the bounded grid-search structure of BGR. For each channel, we generate a set of candidate clipping intervals $[l_c^{(r)},r_c^{(r)}]$ by progressively shrinking the original range $[\min_j w_{c,j},\max_j w_{c,j}]$. For each candidate interval, we initialize an asymmetric affine quantizer and alternately update the scale and zero point. Given the current integer code $q_{c,j}$, the weighted scale update is

\begin{equation}
    s_c \leftarrow
    \frac{\sum_j \rho_{c,j}(q_{c,j}-z_c)w_{c,j}}
    {\sum_j \rho_{c,j}(q_{c,j}-z_c)^2},
    \label{eq:tbgr_scale_update}
\end{equation}
followed by the weighted zero-point update
\begin{equation}
    z_c \leftarrow
    \operatorname{clip}\left(
    \left\lfloor
    \frac{\sum_j\rho_{c,j}\left(q_{c,j}-w_{c,j}/s_c\right)}
    {\sum_j\rho_{c,j}}
    \right\rceil,
    0,2^{b_w}-1
    \right).
    \label{eq:tbgr_zero_point_update}
\end{equation}
The codes are then requantized under the updated $(s_c,z_c)$, and the refined parameters with the smallest value of Eq.~\eqref{eq:tbgr_loss} are selected.

\subsection{Coarse-to-Fine Activation Precision Scheduling}
\label{sec:caps}

\subsubsection{Motivation}

Uniform activation precision does not account for position-dependent sensitivity in coarse-to-fine autoregressive generation. Early stages contain relatively few tokens but establish the global structural scaffold on which all subsequent stages condition. Later stages operate over denser token grids and mainly refine local content. This ordering gives a practical prior for precision allocation: early perturbations can affect the context consumed by later stages, whereas later perturbations are more localized. CAPS uses this prior by reserving the limited high-precision budget for an early prefix of the autoregressive trajectory.

\subsubsection{Stage-wise Linear FLOPs Proxy}

For a quantized linear projection $m$ with input and output dimensions $K_m$ and $N_m$, processing $n$ tokens requires the linear FLOPs proxy $2nK_mN_m$. At stage $i$, the scale vector $\mathbf{p}_i$ defines a $D$-dimensional token grid, and $T_i=\prod_{d=1}^{D}p_{i,d}$ is the number of image tokens in that stage. Thus, image-token projections process $BT_i$ tokens for batch size $B$. We use $F_i$ as the linear-layer FLOPs proxy for precision allocation. We also include text-conditioning KV projections using the number of conditional tokens $T_{\mathrm{cond}}$. 
Let $n_{i,m}$ denote the resulting token count associated with projection $m$ at stage $i$; the stage cost is

\begin{equation}
    \begin{aligned}
        F_i &= \sum_{m\in\mathcal{M}} 2n_{i,m}K_mN_m, \\
        n_{i,m} &=
        \begin{cases}
            BT_i, & m\ \text{is an image-token projection},\\
            T_{\mathrm{cond}}, & m\ \text{is a text-KV projection}.
        \end{cases}
    \end{aligned}
    \label{eq:caps_flops}
\end{equation}

\subsubsection{FLOPs-Budgeted Precision Allocation}

CAPS assigns an activation bit-width $b_i\in\{b_{\mathrm{low}},b_{\mathrm{high}}\}$ to each autoregressive stage while enforcing a FLOPs-weighted average-bit constraint:
\begin{equation}
    \bar{b}_{\mathrm{act}}
    =\frac{\sum_{i=1}^{S}F_i b_i}{\sum_{i=1}^{S}F_i}
    \leq B_{\mathrm{target}}.
    \label{eq:caps_average_bits}
\end{equation}
Equivalently, the FLOPs that can be assigned high precision are bounded by
\begin{equation}
    F_{\mathrm{high}}^{\mathrm{budget}}
    =
    \left\lfloor
    \frac{(B_{\mathrm{target}}-b_{\mathrm{low}})\sum_{i=1}^{S}F_i}
    {b_{\mathrm{high}}-b_{\mathrm{low}}}
    \right\rfloor.
    \label{eq:caps_high_budget}
\end{equation}
The choice of $b_{\mathrm{low}}$, $b_{\mathrm{high}}$, and $B_{\mathrm{target}}$ specifies the available precision budget. Because stages are indivisible allocation units, the realized $\bar{b}_{\mathrm{act}}$ can be lower than the target.

\subsubsection{Prefix-Greedy Scheduling and Runtime Switching}

To instantiate the coarse-to-fine prior, CAPS uses a prefix-greedy allocation. Starting from the earliest stage, it assigns $b_{\mathrm{high}}$ until the next stage would exceed Eq.~\eqref{eq:caps_high_budget}; all remaining stages use $b_{\mathrm{low}}$. Formally, the high-precision set is a prefix $\mathcal{H}=\{1,\ldots,k\}$, where $k$ is the largest index satisfying

\begin{equation}
    \sum_{i=1}^{k}F_i\leq F_{\mathrm{high}}^{\mathrm{budget}}.
    \label{eq:caps_prefix}
\end{equation}

At inference time, CAPS constructs the plan once from the input scale schedule and switches the activation bit-width of all targeted quantized linear projections at each stage boundary. The prefill computation is initialized at $b_{\mathrm{high}}$ to protect the initial conditioning and structural context; Eq.~\eqref{eq:caps_average_bits} is defined over the scheduled autoregressive coarse-to-fine stages. This runtime stage-adaptive behavior uses the same quantized linear modules throughout generation, while their activation precision changes as the generation stage progresses.

\begin{algorithm}[t]

\caption{HGA-Quant pipeline for autoregressive visual generation.}
\label{alg:hga_quant}
\begin{algorithmic}[1]
\Require pretrained ARVG model $f_{\theta}$; target modules $\mathcal{M}$; scale schedule $\mathcal{S}$; quantizer settings $\mathcal{C}$.
\Ensure quantized model $f_{\widehat{\theta}}$ and stage-wise activation plan $\Pi$.
\State $\mathcal{M}_q\gets\varnothing$
\State apply the fixed random Hadamard representation transform when used by the target implementation
\State \textbf{Offline weight calibration:}
\For{each target module $m\in\mathcal{M}$ with weight $W_m$}
    \State $(Q_m,s_m,z_m)\gets\Call{T-BGR}{W_m,\mathcal{C}}$
    \State replace $m$ with a quantized linear module storing $(Q_m,s_m,z_m)$
    \State $\mathcal{M}_q\gets\mathcal{M}_q\cup\{m\}$
\EndFor
\State \textbf{Runtime activation scheduling:}
\State $\Pi\gets\Call{CAPS}{\mathcal{S},\mathcal{M}_q,\mathcal{C}}$
\State run prefill with high-bit activations
\For{stage $i=1,\ldots,|\mathcal{S}|$}
    \State set all modules in $\mathcal{M}_q$ to activation bit-width $\Pi_i$
    \State generate the tokens of stage $i$
\EndFor
\State \Return $f_{\widehat{\theta}},\Pi$
\end{algorithmic}
\end{algorithm}


\providecommand{\hgaBest}[1]{\textcolor{red}{\textbf{#1}}}
\providecommand{\hgaSecond}[1]{\textcolor{blue}{#1}}

\section{Experiments}
\label{sec:experiments}

\subsection{Experimental Setup}
\label{sec:experimental_setup}

\paragraph{Models and Evaluation Protocols.}
We evaluate HGA-Quant on four ARVG backbones: Infinity-2B~\citep{Infinity}, HART~\citep{HART}, VAR-d30~\citep{VAR}, and FlexVAR~\citep{FlexVAR}. For Infinity, our primary testbed, we evaluate text-image alignment using GenEval~\citep{GenEval} and DPG-Bench~\citep{ELLA-DPG}, and assess preference-oriented visual quality using ImageReward~\citep{ImageReward}. For HART, we evaluate prompt adherence using DPG-Bench and visual quality using Fr\'echet Inception Distance (FID) on the MJHQ-30K benchmark~\citep{MJHQ-30K}. For VAR-d30 and FlexVAR, we generate 50K images on ImageNet~\citep{ImageNet} and evaluate generation quality using FID~\citep{FID} and Inception Score (IS)~\citep{IS}. All experiments run on a single NVIDIA RTX A6000 GPU for image generation and evaluation.

\paragraph{Compared Methods.}
Baselines include RTN (Round-to-Nearest), SmoothQuant~\citep{SmoothQuant}, SVDQuant~\citep{SVDQuant}, QuaRot~\citep{QuaRot}, and PTQ4ARVG~\citep{PTQ4ARVG}; PTQ4ARVG uses 128 MS COCO captions for calibration~\citep{COCO}.

\paragraph{Configurations.}
We denote $b$-bit weights and $a$-bit activations by $W_bA_a$. \emph{Ours} is uniform-precision T-BGR with $\kappa=3,\gamma=8$; \emph{Ours (CAPS)} uses the same weight calibration with FLOPs-budgeted A8/A4 stage scheduling targeting W4A6, whose realized average may fall below 6 bits because stages are indivisible. Our method and QuaRot use fixed Hadamard preprocessing; other baselines use their standard settings. Red/blue mark the best/second-best quantized results.

\subsection{Main Results}
\label{sec:main_results}

\begin{table*}[!t]
\vspace{-10mm}
\caption{Unified comparison on Infinity-2B and HART.}

\label{tab:infinity_hart_results}
\centering
\scriptsize
\setlength{\tabcolsep}{2.8pt}
\renewcommand{\arraystretch}{1.08}
\begin{tabular*}{\linewidth}{@{\extracolsep{\fill}}clccccc@{}}
\toprule
\multirow{2}{*}{Quantization} &
\multirow{2}{*}{Methods} &
\multicolumn{3}{c}{\textbf{Infinity-2B}} &
\multicolumn{2}{c}{\textbf{HART}} \\
\cmidrule(lr){3-5}
\cmidrule(l){6-7}
& &
\textbf{GenEval} $\uparrow$ &
\textbf{ImageReward} $\uparrow$ &
\textbf{DPG-Bench} $\uparrow$ &
\textbf{DPG-Bench} $\uparrow$ &
\shortstack{\textbf{MJHQ-30K}\\\textbf{FID} $\downarrow$} \\
\midrule
FP & -- &
0.71896 & 0.95485 & 83.4741 &
74.835 & 6.178 \\
\midrule
\multirow{6}{*}{W8A8}
& RTN &
0.72706 & 0.90395 & 83.0017 &
\hgaBest{75.754} & 12.476 \\
& SmoothQuant &
0.72192 & 0.92828 & 83.0177 &
75.354 & 7.185 \\
& SVDQuant &
0.71827 & 0.93168 & 82.8461 &
\hgaSecond{75.411} & 7.477 \\
& QuaRot &
0.72866 & \hgaSecond{0.94160} & \hgaBest{83.2944} &
75.019 & \hgaSecond{6.380} \\
& PTQ4ARVG &
\hgaBest{0.74081} & 0.89858 & 82.971 &
-- & -- \\
& Ours &
\hgaSecond{0.73216} & \hgaBest{0.94418} & \hgaSecond{83.1832} &
75.172 & \hgaBest{6.366} \\
\midrule
\multirow{6}{*}{W4A8}
& RTN &
0.61900 & 0.33108 & 79.1849 &
62.029 & 92.908 \\
& SmoothQuant &
0.51707 & -0.14869 & 75.0448 &
49.257 & 117.259 \\
& SVDQuant &
0.61737 & 0.26062 & 78.9916 &
\hgaBest{72.656} & 26.802 \\
& QuaRot &
0.66563 & \hgaSecond{0.75864} & 80.9206 &
\hgaSecond{72.173} & \hgaSecond{16.669} \\
& PTQ4ARVG &
\hgaSecond{0.68797} & 0.64962 & \hgaSecond{81.6667} &
-- & -- \\
& Ours &
\hgaBest{0.69825} & \hgaBest{0.83010} & \hgaBest{82.0691} &
71.895 & \hgaBest{9.379} \\
\midrule
\multirow{7}{*}{W4A6}
& RTN &
0.44685 & -0.29458 & 71.0519 &
17.199 & 186.260 \\
& SmoothQuant &
0.28231 & -0.93134 & 66.0039 &
18.593 & 145.355 \\
& SVDQuant &
0.48178 & -0.30246 & 75.5081 &
64.644 & 82.087 \\
& QuaRot &
0.65560 & 0.76999 & 80.8647 &
\hgaSecond{71.386} & 23.916 \\
& PTQ4ARVG &
0.65213 & 0.50648 & 81.2235 &
-- & -- \\
& Ours &
\hgaSecond{0.68003} & \hgaSecond{0.82882} & \hgaSecond{82.0022} &
\hgaBest{73.211} & \hgaBest{13.166} \\
& Ours (CAPS) &
\hgaBest{0.69043} & \hgaBest{0.83523} & \hgaBest{82.0877} &
71.058 & \hgaSecond{21.675} \\
\midrule
\multirow{6}{*}{W4A4}
& RTN &
0.00312 & -2.25143 & 8.2155 &
5.340 & 307.111 \\
& SmoothQuant &
0.00141 & -2.24901 & 6.0742 &
8.749 & 191.014 \\
& SVDQuant &
0.00469 & -2.25759 & 4.8947 &
10.072 & 195.629 \\
& QuaRot &
\hgaSecond{0.47687} & \hgaSecond{0.25269} & \hgaSecond{75.3323} &
\hgaSecond{50.814} & \hgaSecond{134.535} \\
& PTQ4ARVG &
0.02376 & -2.18336 & 26.2649 &
-- & -- \\
& Ours &
\hgaBest{0.60207} & \hgaBest{0.44772} & \hgaBest{80.1773} &
\hgaBest{61.417} & \hgaBest{88.475} \\
\bottomrule
\end{tabular*}
\end{table*}


\begin{table*}[!t]
\caption{Unified comparison on VAR-d30 and FlexVAR.}
\label{tab:var_flexvar_results}
\centering
\scriptsize
\setlength{\tabcolsep}{3.2pt}
\renewcommand{\arraystretch}{1.08}
\begin{tabular*}{\linewidth}{@{\extracolsep{\fill}}clcccc@{}}
\toprule
\multirow{2}{*}{Quantization} &
\multirow{2}{*}{Methods} &
\multicolumn{2}{c}{\textbf{VAR-d30}} &
\multicolumn{2}{c}{\textbf{FlexVAR}} \\
\cmidrule(lr){3-4}
\cmidrule(l){5-6}
& &
\textbf{FID} $\downarrow$ &
\textbf{IS} $\uparrow$ &
\textbf{FID} $\downarrow$ &
\textbf{IS} $\uparrow$ \\
\midrule
FP & -- &
1.9827 & 323.326 &
2.1598 & 336.262 \\
\midrule
\multirow{6}{*}{W8A8}
& RTN &
1.9742 & 312.803 &
2.2093 & 329.612 \\
& SmoothQuant &
2.0138 & 318.025 &
2.1797 & 331.019 \\
& SVDQuant &
1.9799 & 319.311 &
\hgaSecond{2.1680} & 329.660 \\
& QuaRot &
\hgaSecond{1.9708} & \hgaSecond{322.788} &
\hgaBest{2.1604} & \hgaSecond{335.229} \\
& PTQ4ARVG &
2.4351 & 314.487 &
-- & -- \\
& Ours &
\hgaBest{1.9631} & \hgaBest{323.172} &
2.1854 & \hgaBest{336.288} \\
\midrule
\multirow{6}{*}{W4A8}
& RTN &
54.7231 & 45.315 &
10.4348 & 157.526 \\
& SmoothQuant &
13.1672 & 146.123 &
6.3042 & 209.413 \\
& SVDQuant &
6.6533 & 208.545 &
5.7541 & 229.426 \\
& QuaRot &
\hgaSecond{3.9433} & \hgaSecond{255.786} &
\hgaSecond{3.6883} & \hgaSecond{264.783} \\
& PTQ4ARVG &
26.9862 & 85.808 &
-- & -- \\
& Ours &
\hgaBest{2.5532} & \hgaBest{293.163} &
\hgaBest{2.7165} & \hgaBest{297.213} \\
\midrule
\multirow{7}{*}{W4A6}
& RTN &
92.5568 & 17.685 &
37.4952 & 59.187 \\
& SmoothQuant &
21.2844 & 102.013 &
16.2536 & 130.247 \\
& SVDQuant &
12.4921 & 150.365 &
14.8319 & 146.591 \\
& QuaRot &
4.2578 & 249.087 &
\hgaSecond{4.1719} & \hgaSecond{254.551} \\
& PTQ4ARVG &
37.2724 & 63.484 &
-- & -- \\
& Ours &
\hgaBest{2.6586} & \hgaBest{290.467} &
\hgaBest{2.9150} & \hgaBest{290.874} \\
& Ours (CAPS) &
\hgaSecond{2.9469} & \hgaSecond{283.009} &
8.6066 & 201.977 \\
\midrule
\multirow{6}{*}{W4A4}
& RTN &
195.3049 & 4.439 &
233.0478 & 3.004 \\
& SmoothQuant &
173.3009 & 4.683 &
221.9740 & 3.719 \\
& SVDQuant &
153.7270 & 5.418 &
225.9401 & 3.537 \\
& QuaRot &
\hgaSecond{16.9405} & \hgaSecond{143.519} &
\hgaSecond{44.4773} & \hgaSecond{55.159} \\
& PTQ4ARVG &
109.2316 & 11.218 &
-- & -- \\
& Ours &
\hgaBest{5.9785} & \hgaBest{232.040} &
\hgaBest{17.6975} & \hgaBest{134.295} \\
\bottomrule
\end{tabular*}
\vspace{-5mm}
\end{table*}

\paragraph{Infinity.}

T-BGR is competitive as activation precision decreases: at W4A4 it obtains 0.44772 ImageReward and 80.1773 DPG-Bench, compared with 0.25269 and 75.3323 for QuaRot. Most other methods collapse on at least one metric at W4A4. At the nominal W4A6 target, \emph{Ours (CAPS)} reaches 0.69043, 0.83523, and 82.0877 on GenEval, ImageReward, and DPG-Bench with only 5.5711 FLOPs-weighted activation bits, improving over uniform W4A6 T-BGR (0.68003/0.82882/82.0022).

\paragraph{HART, VAR-d30, and FlexVAR.}

Uniform T-BGR is strongest at W4A6 on these backbones, while CAPS realizes 5.4963, 5.5000, and 5.5000 effective activation bits, respectively, and loses some quality from the lower final budget. On HART, CAPS reaches 71.058 DPG-Bench and 21.675 FID, versus 73.211 and 13.166 for uniform W4A6; the FID is still better than QuaRot's 23.916, although DPG-Bench is slightly lower than QuaRot's 71.386, and both are far better than the other baselines. On VAR-d30, CAPS obtains 2.9469/283.009 (FID/IS), below uniform W4A6's 2.6586/290.467 but better than QuaRot on both metrics. On FlexVAR, CAPS obtains 8.6066/201.977 versus 2.9150/290.874 for uniform W4A6 and 4.1719/254.551 for QuaRot, while remaining better than other baselines. The reduced activation budget can hurt absolute quality, but coarse-to-fine allocation still outperforms most alternatives at lower compute.

\paragraph{CAPS stage sensitivity.}
Lowering one consecutive stage pair from A8 to A4 reduces ImageReward by 0.17448 for the first pair but only 0.00442 for stages 11--12 (Figure~\ref{fig:caps_stage_sensitivity}). CAPS achieves the lowest FLOPs-weighted activation precision (5.5711 bits) and highest ImageReward (0.83523), supporting a high-precision prefix and low-precision refinement.

\newpage
\vspace{-3mm}
\subsection{Ablation Study}
\label{sec:ablation}
\vspace{-2mm}

\begin{wrapfigure}[15]{r}{0.35\linewidth}
    \vspace{-0.3\baselineskip}
    \centering
    \includegraphics[width=\linewidth]{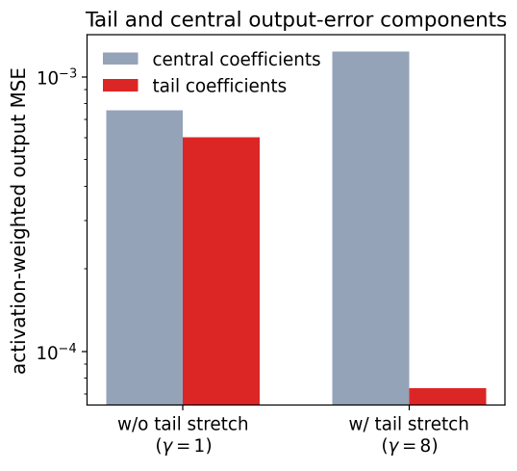}
    \caption{Effect of the tail-stretch factor in T-BGR.}
    \label{fig:tbgr_error_decomposition}
\end{wrapfigure}

\paragraph{T-BGR.}
We investigate the tail-stretch factor $\gamma$ in Tail-Aware Bounded-init Grid Refinement (T-BGR), which controls the contribution of distribution-tail coefficients when optimizing the scale and zero point. Here, $\gamma=1$ corresponds to ordinary BGR, while $\gamma>1$ increases the reconstruction weight of coefficients satisfying $|w-\mu_c|>3\sigma_c$. We evaluate W4A5.6 with coarse-to-fine mixed precision, uniform W4A6, and uniform W4A4 on Infinity-2B using ImageReward. As shown in Table~\ref{tab:ablation}, the main setting $\gamma=8$ improves over no BGR in all three configurations and reaches 0.83523 for W4A5.6 and 0.82882 for W4A6. The optimum is not monotonic in the tail-stretch factor: W4A4 is highest at $\gamma=0$, which assigns zero weight to tail reconstruction error and emphasizes the central mass. This suggests that the most aggressive W4A4 regime favors preserving the dense central region over the rare extremes.

We further isolate the weight-side effect with FP activations and fixed Hadamard preprocessing. Under W4 weight quantization with ordinary BGR ($\gamma=1$), ImageReward is 0.79494. Restoring the detected tail coefficients to FP raises it to 0.82331 when restoring a random half of the tail entries and to 0.83709 when restoring all of them, indicating that a small tail subset has a disproportionate influence on visual quality. T-BGR then tests whether this insight can be used without an FP outlier branch: assigning a larger tail-stretch factor, $\gamma=8$, reaches 0.84753, higher than both ordinary BGR and full tail restoration under $\gamma=1$. This supports treating T-BGR as a range-allocation mechanism that protects tail information while retaining the standard low-bit weight representation.

\begin{table}[H]
\centering
\scriptsize
\vspace{-3mm}
\caption{Ablation studies on Infinity-2B. 
Left: ablation of the T-BGR tail-stretch factor $\gamma$.
Right: ablation of CAPS.
W4A5.6 combines W4A4 and W4A8 at a FLOPs-weighted effective activation precision of 5.5711 bits;
``no BGR'' omits bounded grid refinement.}
\label{tab:ablation}
\vspace{-3mm}

\begin{minipage}[t]{0.50\linewidth}
\setlength{\tabcolsep}{2pt}

\resizebox{\linewidth}{!}{%
\begin{tabular}{lcccccc}
\toprule
 & \textbf{no BGR} & \textbf{$\gamma=0$} & \textbf{$\gamma=1$} & \textbf{$\gamma=2$}  & \textbf{$\gamma=8$} & \textbf{$\gamma=16$} \\
\midrule
W4A5.6
& 0.81594 & 0.81001 & 0.81458 & 0.80069 & \hgaBest{0.83523} & 0.81491 \\
W4A6
& 0.82125 & 0.80062 & 0.80668 & 0.79104 & \hgaBest{0.82882} & 0.79840 \\
W4A4
& 0.40181 & \hgaBest{0.50311} & 0.48880 & 0.49401 & 0.44772 & 0.43064 \\
\bottomrule
\end{tabular}%
}

\end{minipage}
\hspace{0.01\linewidth}
\begin{minipage}[t]{0.45\linewidth}

\centering
\scriptsize

\resizebox{\linewidth}{!}{%
\begin{tabular}{lccc}
\toprule
 & \textbf{GenEval}$\uparrow$ & \textbf{ImageReward}$\uparrow$ & \textbf{DPG-Bench}$\uparrow$ \\
\midrule
W4A8
& 0.69825 & 0.83010 & 82.0691 \\
W4A6
& 0.68003 & 0.82882 & 82.0022 \\
W4A4
& 0.60207 & 0.44772 & 80.1773 \\
W4A5.6
& 0.69043 & 0.83523 & 82.0877 \\
\bottomrule
\end{tabular}%
}

\end{minipage}

\end{table}

\vspace{-7mm}
\paragraph{CAPS.}

\begin{figure}[H]
\centering
\includegraphics[width=\linewidth]{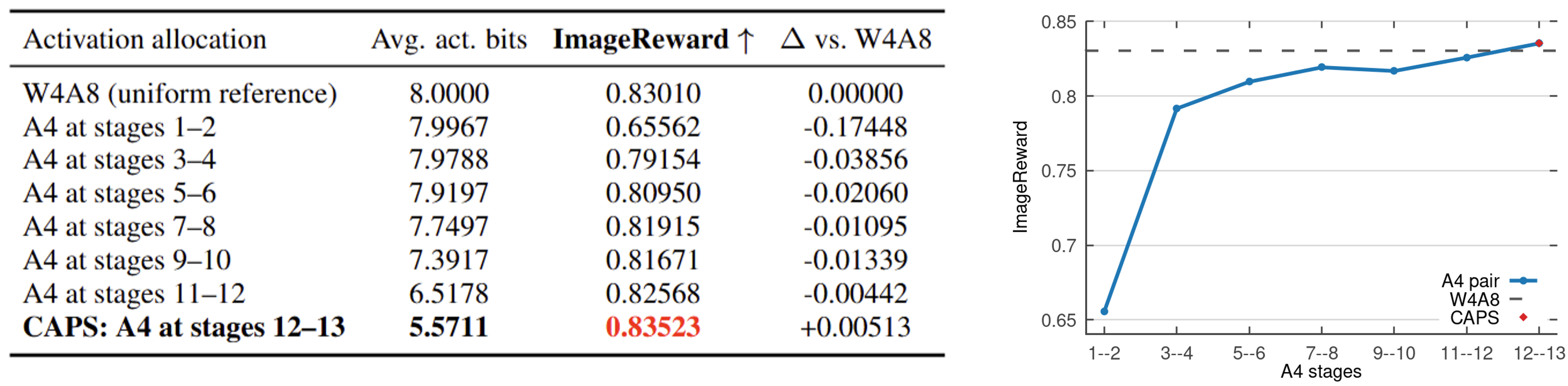}
\vspace{-7mm}
\caption{CAPS stage sensitivity on Infinity-2B. The table reports FLOPs-weighted effective activation bits and ImageReward when changing one consecutive stage pair from A8 to A4; the curve plots the same ImageReward values with the uniform W4A8 reference as a dashed horizontal line.}
\vspace{-6mm}
\label{fig:caps_stage_sensitivity}
\end{figure}

Table~\ref{tab:ablation} evaluates CAPS. W4A5.6 obtains 0.69043, 0.83523, and 82.0877 on GenEval, ImageReward, and DPG-Bench, respectively, compared with 0.68003, 0.82882, and 82.0022 for uniform W4A6. Uniform W4A4 instead drops to 0.60207, 0.44772, and 80.1773. Figure~\ref{fig:caps_stage_sensitivity} motivates prefix protection; Appendix~\ref{app:caps_allocation_comparison} compares early-high, late-high, and random schedules.

\vspace{-2mm}
\section{Conclusion}
\vspace{-2mm}
We presented HGA-Quant, a training-free PTQ framework for low-bit ARVG that combines tail-aware weight calibration with budgeted activation allocation. Experiments across four backbones demonstrate the benefits of T-BGR under aggressive quantization. CAPS supports protecting early coarse stages, but reduced activation budgets can compromise quality on some backbones. These findings motivate backbone-specific precision schedules to balance generation quality and efficiency.

\subsection*{AI use statement}

Generative AI tools were used to assist with drafting and modifying experimental code, including initial implementations of core experimental components and auxiliary scripts. They were also used for language polishing, minor revisions to the organization and content of the manuscript, visualization code, and figure layout ideas.

All AI-assisted code was carefully reviewed, tested, and modified before use, and AI-assisted manuscript content was reviewed and verified before inclusion. The core research ideas, methodology, and experimental design were developed independently of generative AI. Experimental results were analyzed and interpreted by the research team, and the conclusions were independently formulated based on these analyses. The authors take full responsibility for the final content of this work, including all text, code, results, and figures.

\subsection*{Ethics statement}

This work studies post-training quantization methods for improving the efficiency of autoregressive visual generation models. Quantized models may retain the biases, limitations, and potential for misuse of their pretrained counterparts, while reduced computational and memory costs may facilitate broader deployment of such models. Our experiments focus on generation quality and quantization efficiency and do not evaluate the safety, fairness, or potential misuse of the resulting models. The experiments use publicly available models and benchmark datasets and do not involve human subjects or the collection of personal or sensitive data.

\subsection*{Reproducibility statement}

We will publicly release the source code and relevant configurations used to obtain the experimental results reported in this paper. The release will cover our quantization implementations for Infinity, HART, VAR, and FlexVAR, together with the corresponding experimental configurations, to facilitate reproduction of our results.

\bibliography{iclr2027_conference}
\bibliographystyle{iclr2027_conference}

\newpage
\appendix
\section{Algorithm}

\begin{algorithm}[H]
\caption{Tail-Aware Bounded-init Grid Refinement (T-BGR).}
\label{alg:tbgr}
\begin{algorithmic}[1]
\Require weight matrix $W$; bit-width $b_w$; grid size $N$; tail threshold $\kappa$; tail weight $\gamma$; tolerance $\epsilon$; iteration limit $I_{\max}$.
\Ensure integer codes $Q^{\star}$; real-valued scales $s^{\star}$; integer zero points $z^{\star}$.
\State $q_{\max}\gets2^{b_w}-1$
\State compute per-channel statistics $\mu_c,\sigma_c$ of $W_{c,:}$
\State $\rho_{c,j}\gets1+(\gamma-1)\mathbf{1}(|w_{c,j}-\mu_c|>\kappa\sigma_c)$
\State initialize channel-wise best loss $\mathcal{L}^{\star}_c\gets+\infty$
\State $\ell_c\gets\min_j w_{c,j}$, $u_c\gets\max_j w_{c,j}$, and $\eta_c\gets(u_c-\ell_c)/(2N)$
\For{$r=0,\ldots,N-1$}
    \State $W^{(r)}_{c,:}\gets\operatorname{clip}(W_{c,:},\ell_c+r\eta_c,u_c-r\eta_c)$ \Comment{bounded initialization}
    \State initialize $(s,z,Q)$ from the asymmetric affine quantizer induced by $W^{(r)}$
    \For{$t=1,\ldots,I_{\max}$}
        \State $A\gets Q-z$
        \State $s_c\gets
        \dfrac{\sum_j\rho_{c,j}A_{c,j}w_{c,j}}
        {\sum_j\rho_{c,j}A_{c,j}^{2}}$
        \State $z_c\gets
        \operatorname{clip}\!\left(
        \operatorname{round}\!\left(
        \dfrac{\sum_j\rho_{c,j}(q_{c,j}-w_{c,j}/s_c)}
        {\sum_j\rho_{c,j}}\right),
        0,q_{\max}\right)$
        \State $Q\gets\operatorname{clip}(\operatorname{round}(W/s)+z,0,q_{\max})$
        \State $\mathcal{L}_c\gets
        \dfrac{\sum_j\rho_{c,j}\left(w_{c,j}-(q_{c,j}-z_c)s_c\right)^2}
        {\sum_j\rho_{c,j}}$
        \If{$\max_c|\mathcal{L}_c-\mathcal{L}^{\mathrm{prev}}_c|<\epsilon$}
            \State \textbf{break}
        \EndIf
        \State $\mathcal{L}^{\mathrm{prev}}\gets\mathcal{L}$
    \EndFor
    \State update $(Q^{\star}_{c,:},s^{\star}_c,z^{\star}_c,\mathcal{L}^{\star}_c)$ for channels with $\mathcal{L}_c<\mathcal{L}^{\star}_c$
\EndFor
\State \Return $Q^{\star},s^{\star},z^{\star}$
\end{algorithmic}
\end{algorithm}

\begin{algorithm}[H]
\caption{Coarse-to-Fine Activation Precision Scheduling (CAPS).}
\label{alg:caps}
\begin{algorithmic}[1]
\Require scale schedule $\mathcal{S}=\{\mathbf{p}_i\}_{i=1}^{S}$; quantized modules $\mathcal{M}_q$; effective batch size $B$; conditional token count $T_{\mathrm{cond}}$; bit-widths $b_{\mathrm{low}}<b_{\mathrm{high}}$; target average bit-width $b_{\mathrm{tar}}$.
\Ensure stage-wise activation plan $\Pi=(\Pi_1,\ldots,\Pi_S)$.
\For{$i=1,\ldots,S$}
    \State $T_i\gets\prod_d p_{i,d}$
    \State $F_i\gets\sum_{m\in\mathcal{M}_q}2\,n_{i,m}\,d_m^{\mathrm{in}}d_m^{\mathrm{out}}$,
    where $n_{i,m}=T_{\mathrm{cond}}$ for text-KV projections and $n_{i,m}=BT_i$ otherwise
\EndFor
\State $F_{\mathrm{budget}}\gets
\left\lfloor
\dfrac{b_{\mathrm{tar}}-b_{\mathrm{low}}}
{b_{\mathrm{high}}-b_{\mathrm{low}}}
\sum_{i=1}^{S}F_i
\right\rfloor$
\State choose the longest prefix $\mathcal{H}=\{1,\ldots,k\}$ such that $\sum_{i\in\mathcal{H}}F_i\le F_{\mathrm{budget}}$
\For{$i=1,\ldots,S$}
    \State $\Pi_i\gets b_{\mathrm{high}}$ if $i\in\mathcal{H}$, otherwise $\Pi_i\gets b_{\mathrm{low}}$
\EndFor
\State \Return $\Pi$
\end{algorithmic}
\end{algorithm}

\section{Weight-Only Outlier Diagnostic under W3}
\label{app:weight_only_outlier_w3}

The main ablation uses W4 to study tail-aware weight calibration. To check whether the same intuition remains relevant under a more aggressive weight-only setting, we also run W3 quantization with FP activations and fixed Hadamard preprocessing. Without BGR, W3 collapses to -2.25053 ImageReward, while ordinary BGR ($\gamma=1$) recovers the score to 0.58618. This confirms that a range-refinement baseline is important before studying finer tail behavior in very low-bit weights.

Starting from W3 with ordinary BGR, restoring all detected tail coefficients to FP further improves ImageReward from 0.58618 to 0.61887. Restoring only a random half gives 0.57379, slightly below the non-restored run, suggesting that the W3 regime is sensitive to uneven mixed-precision perturbations rather than following a simple monotonic rule. The full-restoration result nevertheless supports the same qualitative conclusion as W4: tail precision remains useful when the bit-width is pushed lower.

We also evaluate the stronger T-BGR setting $\gamma=8$ under W3. It obtains 0.32632, far above the no-BGR collapse but below ordinary BGR. This does not negate the tail-aware motivation; instead, it shows that the appropriate tail emphasis depends on the available bit-width. In an extremely narrow W3 codebook, aggressively stretching the tail can overly reduce the resolution of the central mass, so a bit-width-adaptive tail factor may be preferable.

\section{Coarse-to-Fine Generation Visualization}
\label{app:progressive_generation}

\begin{figure}[H]
    \centering
    \includegraphics[width=\linewidth]{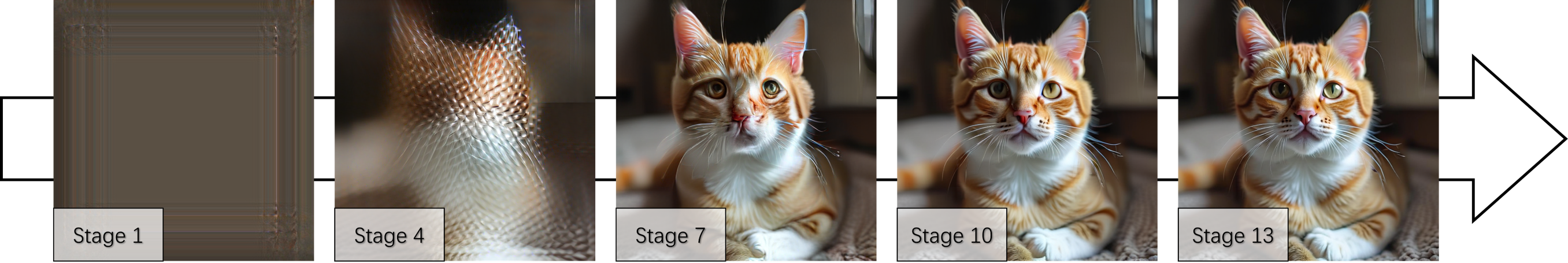}
    \caption{Compact illustration of coarse-to-fine generation at selected stages. The generation process gradually evolves from coarse layouts to detailed images, with early stages determining global structures and later stages refining object-level details.}
    \label{fig:ctf_illustration}
\end{figure}

Figure~\ref{fig:ctf_illustration} provides an intuitive view of the coarse-to-fine generation process of Infinity. The early stages mainly establish global structures and compositions, while later stages progressively refine object details and visual appearance. This observation motivates stage-aware precision allocation: higher precision can be preserved for early structural generation, while lower precision can be applied to later refinement stages.

\begin{figure}[t]
    \centering
    \includegraphics[width=\linewidth]{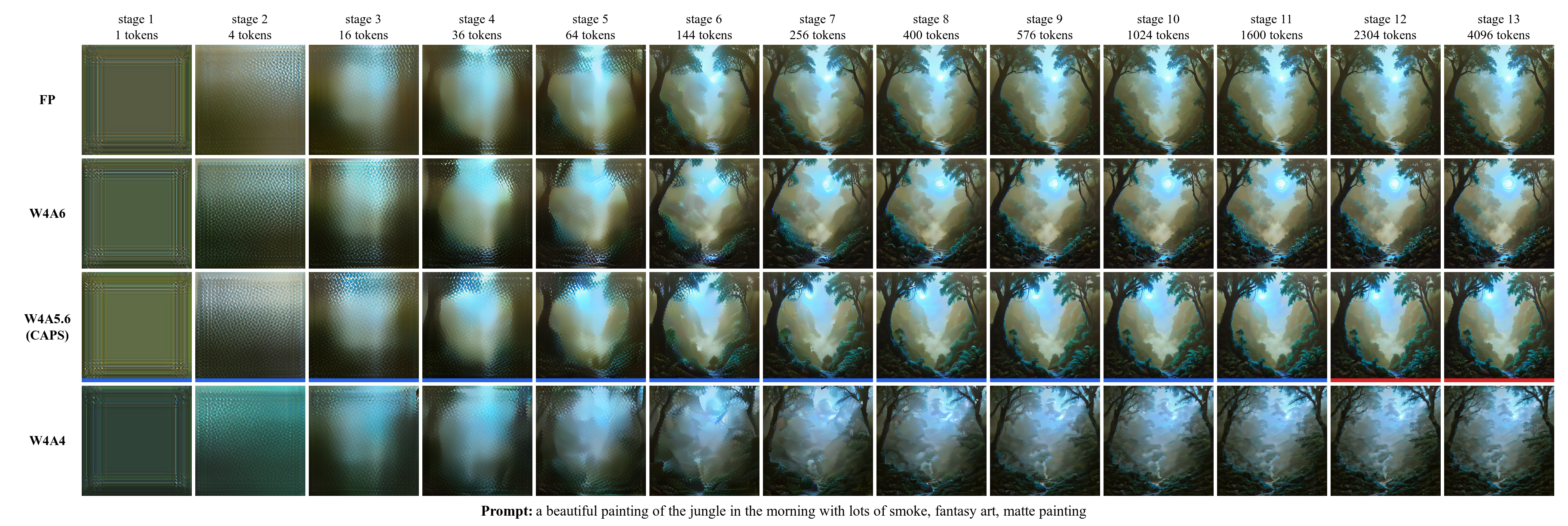}
    \caption{Stage-wise visualization of coarse-to-fine generation. We compare full precision (FP), uniform W4A6, CAPS with W4A5.6, and uniform W4A4 over all 13 generation stages. CAPS preserves the coarse structure with its high-precision prefix and maintains coherent refinement after switching to lower precision in the final stages, whereas uniform W4A4 exhibits visible artifacts throughout the trajectory.}
    \label{fig:progressive_generation}
\end{figure}

To further illustrate this behavior, Figure~\ref{fig:progressive_generation} compares intermediate outputs at all 13 stages under full precision (FP), uniform W4A6, CAPS with W4A5.6, and uniform W4A4. The early stages preserve the global layout, while later stages mainly contribute incremental refinements. CAPS maintains a coherent trajectory after switching to lower precision in the final stages, whereas uniform W4A4 introduces artifacts from early stages and produces a visibly degraded trajectory.

\section{Budgeted Allocation Comparison}
\label{app:caps_allocation_comparison}

The stage-wise results in the main text establish that early and late stages have different precision sensitivity. We therefore add a broader placement check under the same target FLOPs budget. The early-prefix schedule is the CAPS allocation used by our method, with A8 assigned to stages 1--11 and A4 assigned to stages 12--13. The late-high schedule assigns A8 only to the final refinement stage, which has the largest token grid. For random schedules, we start from all-A4 stages, sample a deterministic random stage order, and switch a stage to A8 when doing so does not exceed the target budget; if a stage would exceed the budget, it is skipped and the remaining stages are considered. All rows use the common allocation-comparison entry; consequently, its early-high score is 0.83523.

\begin{table}[H]
\centering
\setlength{\tabcolsep}{4pt}
\caption{Budgeted CAPS allocation comparison on Infinity-2B. Stage labels are one-based. The reported average bit-width is the FLOPs-weighted effective activation precision realized by each allocation, and all rows use the common allocation-comparison evaluation entry.}
\label{tab:caps_allocation_comparison}
\begin{tabular}{lcccc}
\toprule
Allocation & A8 stages & Avg. act. bits & \textbf{ImageReward} $\uparrow$ & $\Delta$ vs.\ early-high \\
\midrule
early-high (CAPS) & 1--11 & 5.5711 & \textbf{0.83523} & 0.00000 \\
late-high & 13 & 5.5543 & 0.44577 & -0.38946 \\
\texttt{random\_01} & 1,2,3,4,5,6,8,9,11,12 & 5.9587 & 0.81186 & -0.02337 \\
\texttt{random\_02} & 1,2,3,4,5,10,13 & 5.9930 & 0.77997 & -0.05526 \\
\texttt{random\_03} & 1,2,3,4,5,6,7,8,9,10,12 & 5.8381 & 0.82448 & -0.01075 \\
\texttt{random\_04} & 1,2,3,4,5,6,7,8,13 & 5.9095 & 0.80030 & -0.03493 \\
\texttt{random\_05} & 1,2,3,4,5,8,9,13 & 5.9755 & 0.78069 & -0.05454 \\
\bottomrule
\end{tabular}
\end{table}

The early-high and late-high schedules are also close in compute: their high-precision linear FLOPs are 31.20T and 30.87T, differing by 1.06\% of the high-precision portions and only 0.42\% of the full stage FLOPs. Nevertheless, ImageReward falls from 0.83523 to 0.44577, a decrease of 0.38946 that is close to uniform W4A4 (0.44772). Among the random schedules, the best result is \texttt{random\_03} at 0.82448, still 0.01075 below early-high despite using a higher realized activation precision of 5.8381 bits. Thus, neither a similar compute budget nor a larger average precision is sufficient without protecting the early stages.

\section{Qualitative Comparisons}

We provide qualitative comparisons across the evaluated methods under W4A6 and coarse-to-fine W4A5.6 quantization on Infinity-2B. For each example, all methods are conditioned on the same text prompt. Our method generally preserves the global composition and fine-grained details more faithfully, while several baselines exhibit visible artifacts, structural distortion, or severe quality degradation at aggressive bit-widths.

\begin{figure}[!h]
    \centering
    \includegraphics[width=\linewidth]{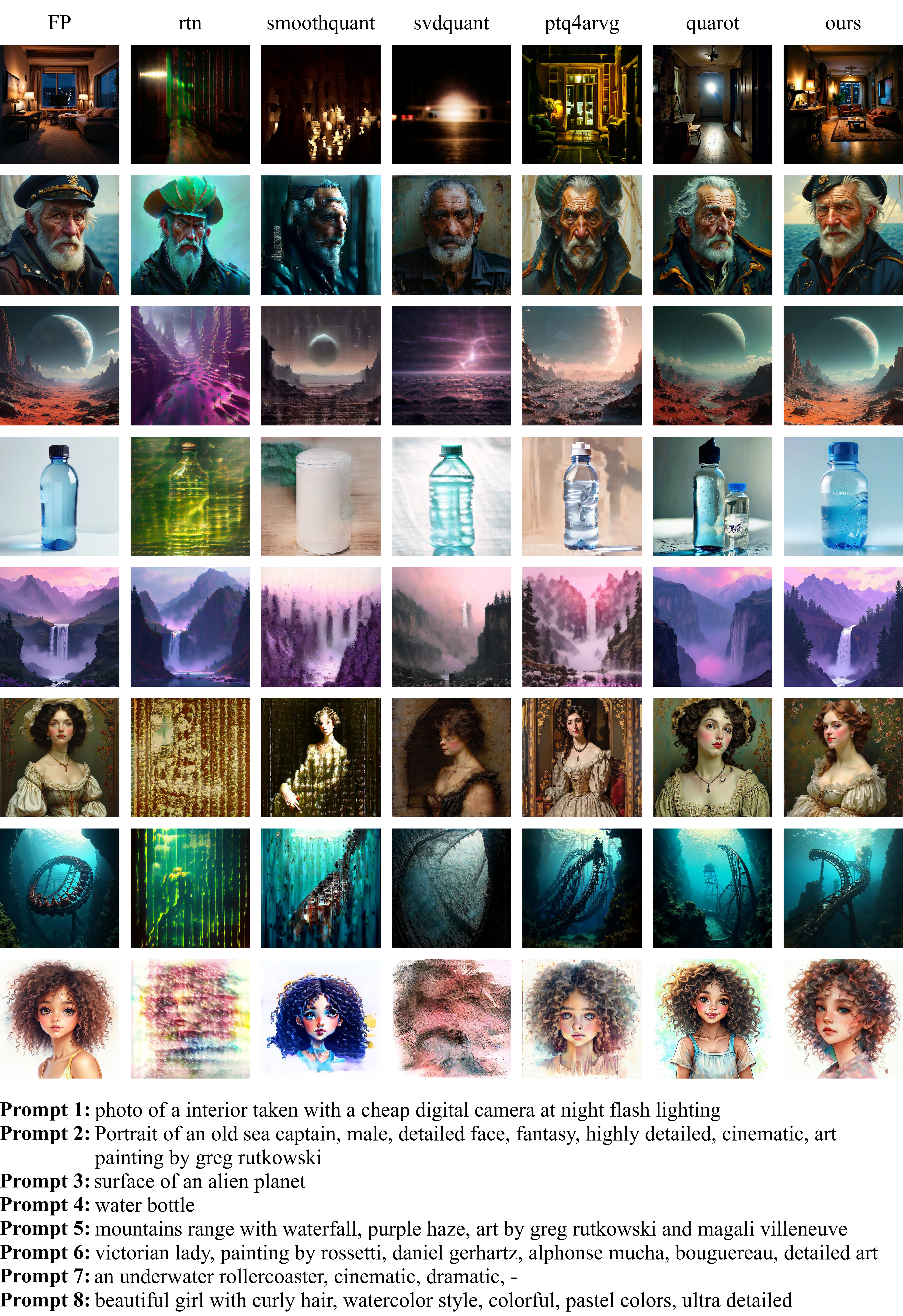}
    \caption{Comparisons across different methods. In this figure, all images were obtained with the model undergoing W4A6 quantization.}
    \label{fig:last_w4a6}
\end{figure}

\begin{figure}[!h]
    \centering
    \includegraphics[width=\linewidth]{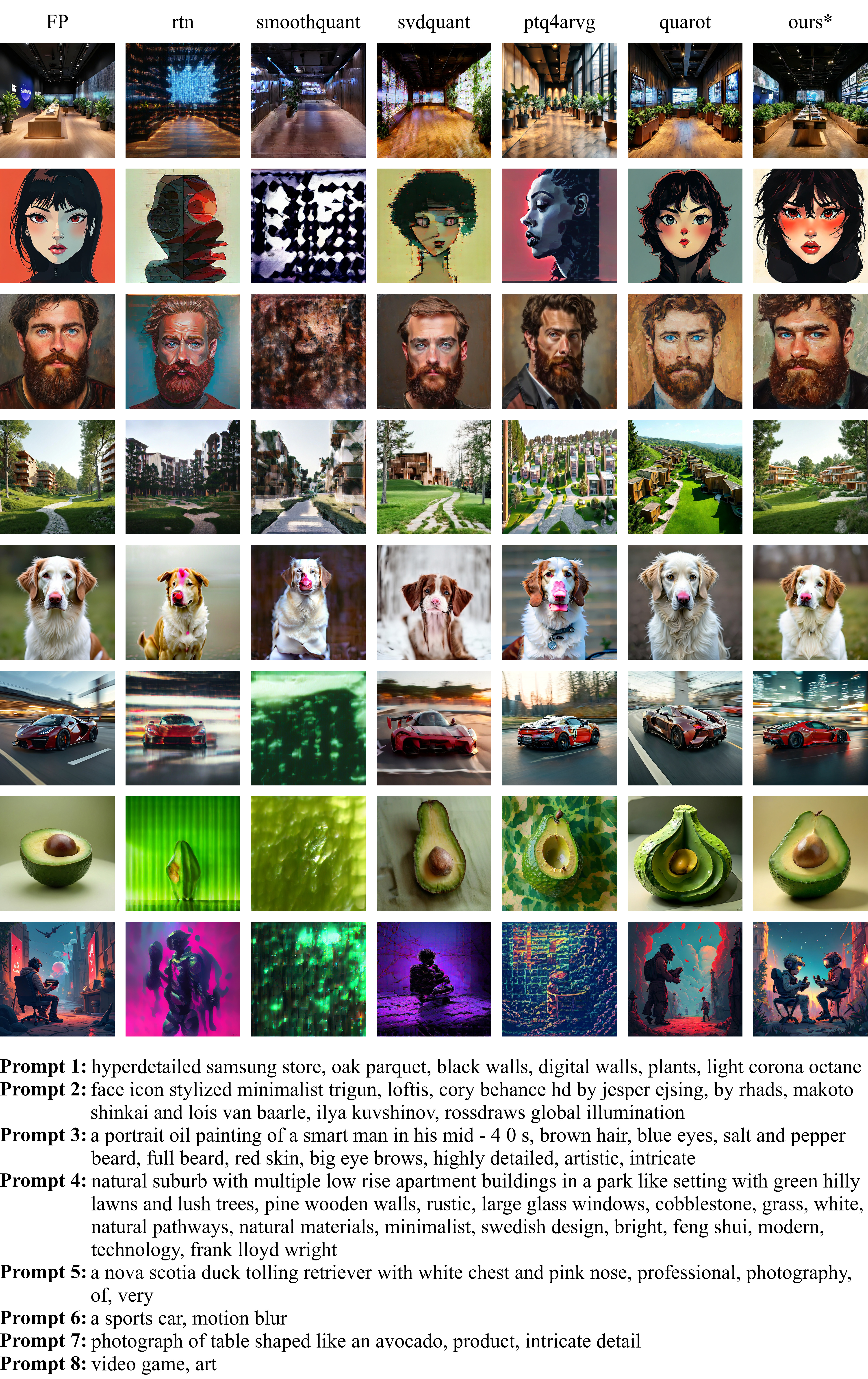}
    \caption{In this figure, the images labeled "ours*" were obtained using our coarse-to-fine W4A5.6 approach, while the images from other methods were generated using W4A6 quantization.}
    \label{fig:last_ctf}
\end{figure}

\end{document}